\documentclass[12pt]{elsarticle}
\makeatletter
\def\ps@pprintTitle{%
 \let\@oddhead\@empty
 \let\@evenhead\@empty
 \def\@oddfoot{}%
 \let\@evenfoot\@oddfoot}
\makeatother

\usepackage{graphicx}
\usepackage{amssymb,bm,mathtools}
\usepackage{amsthm}
\usepackage{cancel}
\usepackage{listings}
\usepackage{color}
\definecolor{codegreen}{rgb}{0,0.6,0}
\definecolor{codegray}{rgb}{0.5,0.5,0.5}
\definecolor{codepurple}{rgb}{0.58,0,0.82}
\definecolor{backcolour}{rgb}{0.95,0.95,0.92}

\lstdefinestyle{mystyle}{
  backgroundcolor=\color{backcolour},   commentstyle=\color{codegreen},
  keywordstyle=\color{magenta},
  numberstyle=\tiny\color{codegray},
  stringstyle=\color{codepurple},
  basicstyle=\footnotesize,
  breakatwhitespace=false,         
  breaklines=true,                 
  captionpos=b,                    
  keepspaces=true,                 
  numbers=left,                    
  numbersep=5pt,                  
  showspaces=false,                
  showstringspaces=false,
  showtabs=false,                  
  tabsize=2
}

\usepackage{hyperref}
\hypersetup{
    colorlinks=true,
    linkcolor=blue,
    filecolor=magenta,      
    urlcolor=cyan,
}
\usepackage{xargs}                      
\usepackage[pdftex,dvipsnames]{xcolor}  
\usepackage[colorinlistoftodos,prependcaption,textsize=tiny]{todonotes}
\newcommandx{\unsure}[2][1=]{\todo[linecolor=red,backgroundcolor=red!25,bordercolor=red,#1]{#2}}
\newcommandx{\change}[2][1=]{\todo[linecolor=blue,backgroundcolor=blue!25,bordercolor=blue,#1]{#2}}
\newcommandx{\info}[2][1=]{\todo[linecolor=OliveGreen,backgroundcolor=OliveGreen!25,bordercolor=OliveGreen,#1]{#2}}
\newcommandx{\improvement}[2][1=]{\todo[linecolor=Plum,backgroundcolor=Plum!25,bordercolor=Plum,#1]{#2}}
\newcommandx{\thiswillnotshow}[2][1=]{\todo[disable,#1]{#2}}
\usepackage{lineno}
\theoremstyle{definition}
\newtheorem{definition}{Definition}[section]
\newtheorem{theorem}{Theorem}[section]
\newtheorem{corollary}{Corollary}[theorem]
\newtheorem{exmp}{Example}[section]
\newtheorem{lemma}[theorem]{Lemma}
\DeclareMathOperator{\tr}{tr} 
\DeclareMathOperator{\E}{\mathbb{E}}
\usepackage{tikz,standalone}
\usetikzlibrary{positioning,calc,shapes.geometric, arrows}
\begin{document}

\begin{frontmatter}
\title{Hands-on Experience with Gaussian Processes (GPs): Implementing GPs in Python - I}

\author[aalto]{Kshitij Tiwari\corref{cor1}}
\ead{kshitij.tiwari@aalto.fi}
\ead[url]{https://sites.google.com/view/kshitijtiwari/}
\cortext[cor1]{Corresponding author}
\address[aalto]{Intelligent Robotics Group, Department of Electrical Engg. \& Automation, Aalto University, Espoo, 02150, Finland}

\begin{abstract}
This document serves to complement our \href{https://sites.google.com/view/exposure-to-gp/}{website} which was developed with the aim of exposing the students to Gaussian Processes (GPs). GPs are non-parametric bayesian regression models that are largely used by statisticians and geospatial data scientists for modeling spatial data. Several open source libraries spanning from Matlab \cite{rasmussen2010gaussian}, Python \cite{gpy2014}, R \cite{laGP}  \textit{etc.} are already available for simple plug-and-use. The objective of this handout and in turn the website was to allow the users to develop stand-alone GPs in Python by relying on minimal external dependencies. To this end, we only use the default python modules and assist the users in developing their own GPs from scratch giving them an in-depth knowledge of what goes on under the hood. The module covers GP inference using maximum likelihood estimation (MLE) and gives examples for 1D (dummy) spatial data.
\end{abstract}

\end{frontmatter}


\cleardoublepage
\tableofcontents
\lstlistoflistings
\cleardoublepage

\section{Gaussian Processes (GPs)}
Gaussian Processes (GPs) were introduced by Carl E. Rasmussen in \cite{rasmussen2004gaussian} and since then have undergone significant development. Formally speaking, GPs are a collection of random variables, a finite collection of which is a multivariate normal distribution. Although it seems like GPs are infinite dimensional entities, but, we almost never have to deal with infinite dimensions at any time. The reason being that we observe a finite-dimensional subset of infinite-dimensional data, and this finite subset follows a multivariate normal distribution.

\subsection{Comparison to Parametric Linear Regression}
In a \textit{linear regression} problem, we try to fit a linear model to explain the relationship between the output variable $\mathbf{y}$ and the input variable $\mathbf{x}$. In generic terms, it can be modeled as $\mathbf{y} = f(\mathbf{x}) + \bm\epsilon$ where $\bm\epsilon$ is the irreducible reconstruction error. Since it is known \textit{a priori} that the relationship is linear, the model $f(\mathbf{x})$ can simply be replaced by a straight line modeled by the intercept parameter $\theta_0$ and slope parameter $\theta_1$ such that $y = \theta_0 + \theta_1 \mathbf{x} + \bm\epsilon$. Then, the problem simply remains to fit this model to the data to infer the values of $\theta_0,\theta_1$. This model then becomes \textit{parametric}.
\begin{figure}[!htbp]
\centering
\includegraphics[scale=0.5]{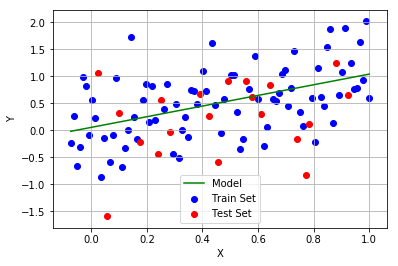}
\caption{Parametric Linear Regression.}
\end{figure}

As opposed to this, in GPs, there is no assumption about the functional form of the model that fits the data. As such, a probabilistic prior is placed over all possible models like linear, exponential \textit{etc.,} and a posterior is obtained to best fit the data. This approach then is \textit{non-parametric}.

\subsection{Notational Conventions}
Let $X$ represent the set of inputs and $Y$ represent the corresponding targets. Then, $X^*$ would represent the unobserved inputs. The corresponding targets would then be represented by $Y^*$ and have to be predicted using the GP posterior. The covariance kernel is represented by $\mathcal{K}(\cdot,\cdot)$ while $[\mathcal{K}]_{ij} \triangleq k(x_i,x_j)$. Let $\bm{\mu}_{f|D}$ represent the posterior mean vector over $X^*$ and $\mathcal{K}_{f|D}$ represent the corresponding posterior covariance matrix. The hyperparameters are denoted by $\theta$. Just like the linear regression case, here it is assumed that $\mathbf{y} = f(\mathbf{x}) + \bm\epsilon$ where $f(\mathbf{x}) \sim \mathcal{GP}(\mu(\cdot),\mathcal{K}(\cdot,\cdot))$ and $\bm\epsilon \sim \mathcal{N}(0,\sigma_n)$.

\subsection{Prior Mean and Covariance Functions}
Without loss of generality, it is often assumed that the GPs have a prior mean of zero \cite{rasmussen2004gaussian}. However, if for some situations this is not applicable, then the problem can simply be addressed by a meager change of variables\footnote{Since this is beyond the scope of the current handout, further details have been omitted.}. Formally speaking, the prior mean for an input $x \in \mathcal{R}^1$ would be defined as:
\begin{equation}
\mu(x) \triangleq \E(f(x))
\end{equation}

As for the covariance, often times people know some information about how the correlations in the spatial data of interest vary. One of the most popular tool is the squared exponential kernel or the RBF kernel which explains the similarity between the targets (outputs) in terms of inverse squared law of spatial separation between the inputs. Mathematically, the squared exponential correlation between inputs $x,x'$ is given by:

\begin{align}
\begin{split}
k(x,x') &\triangleq Cov (f(x),f(x')) \,,\\
& = \E(\{f(x)-\E(f(x))\}\{f(x)-\E(f(x))\}) \,,\\
& = \exp(-||x-x'||^2)
\end{split}
\label{eq:sq_exp_cov}
\end{align}
It must be pointed out here that the kernel is defined only in terms of the spatial separation like that in Eq.~\eqref{eq:sq_exp_cov} is called \textit{stationary} kernel. Such kernels only depend on separation and not the absolute values of $x,x'$ which means that where ever in the domain, the spatial separation is identical, the covariance will be identical. Additionally, the correlation between inputs decays inversely as a function of distance, \textit{i.e.,} closer inputs are highly correlated as compared to farther inputs. 

\subsection{Posterior Mean and Covariance Functions}
Below we begin by considering a simple case of noise free observations and then extend it to noisy observation case.
\subsubsection{Noise Free Case}
The joint distribution of the observed and unobserved inputs $X,X^*$ respectively, is given by:
\begin{gather}
 \begin{bmatrix} \mathbf{f} \\ \mathbf{f^*} \end{bmatrix}
 \sim \mathcal{N}\left(
   \mathbf{0} ,
   \begin{bmatrix}
   \mathcal{K}(X,X) & \mathcal{K}(X,X^*)\\
   \mathcal{K}(X,X^*)^T & \mathcal{K}(X^*,X^*)
   \end{bmatrix}\right)
\end{gather}
where $ \mathcal{K}(X,X)$ represents the auto-correlation between the inputs $X$ and $\mathcal{K}(X,X^*)$ represents the cross-correlations between the observed and unobserved inputs. Similarly, $\mathcal{K}(X^*,X^*)$ represents auto-correlation amongst the unobserved inputs $X^*$. Shorthand for all the aforementioned kernels are $K,K^*,K^{**}$ in the respective order. These notations confirm with the coding exercise which are presented later on.

Now, in order to restrain the posterior distribution to only the functions which agree with the data, we can restrain the posterior possibilities by conditioning on the observations. This gives the posterior distribution as $p(\mathbf{f}^*|X^*,X,\mathbf{f})\sim \mathcal{N}(\bm{\mu}_{f|D},\mathcal{K}_{f|D})$ where $D = [X,Y]$ represents the training dataset and the posterior mean and covariance are explained in Eq.~\eqref{eq:posterior_mean}, Eq.~\eqref{eq:posterior_cov} respectively.
\begin{equation}
\bm\mu_{f|D} \triangleq \underbrace{\cancelto{0}{\bm\mu}^{*}}_\text{Prior} + \underbrace{\mathcal{K^*}^T\mathcal{K}^{-1}(\mathbf{f} - \bm\mu)}_\text{Corrector}
\label{eq:posterior_mean}
\end{equation}
Here, ${\bm\mu}^{*}$ represents the prior mean while $\bm\mu$ represents the mean of the noise free observations $f$. This equation can be intuitively interpreted as a correction to the prior mean by a corrector term which represents the weighted combination of kernel functions summed over each training sample $x\in X$. Eq.~\eqref{eq:posterior_mean} can be seen a linear estimator with $\bm\mu_{f|D} = \mathcal{K^*}^T \bm\alpha$ for $\bm\alpha = \mathcal{K}^{-1}(\mathbf{f} - \bm\mu)$ and this infact, is the best linear unbiased estimator (BLUP).

\begin{equation}
\mathcal{K}_{f|D} \triangleq \underbrace{\underbrace{\mathcal{K^{**}}}_\text{Prior} - \underbrace{\mathcal{K^*}^T\mathcal{K}^{-1}\mathcal{K^*}}_\text{Evidence}}_\text{Reduction in Variance}
\label{eq:posterior_cov}
\end{equation}

This equation clearly shows the reduction in variance as more evidence is acquired from observations. From Eq.~\eqref{eq:posterior_cov}, it is evident that the posterior covariance does not depend on the observations which is the case of Eq.~\eqref{eq:posterior_mean}. However, it must be noted that there is an indirect dependence on the observations since the hyper-parameters of the kernel encode the relationships from observations. For a proof-sketch, refer to Theorem \ref{app:Posterior_Proof} in the Appendix.

\subsubsection{Noisy observation case}
When the observations are noisy, which is usually the case in real-world, the joint posterior must consider the noisy observations. Then, the revised joint posterior is given by:
\begin{gather}
 \begin{bmatrix} Y \\ \mathbf{f^*} \end{bmatrix}
 \sim \mathcal{N}\left(
   \mathbf{0} ,
   \begin{bmatrix}
   \mathcal{K}\textcolor{red}{+\sigma^2_nI} & \mathcal{K^*}\\
   \mathcal{K^*}^T & \mathcal{K^{**}}
   \end{bmatrix}\right)
\end{gather}

The predictive equations are now given by:
\begin{align}
\begin{split}
\bm\mu_{Y|D} \triangleq \underbrace{\cancelto{0}{\bm\mu}^{*}}_\text{Prior} + \mathcal{K^*}^T[\mathcal{K} \textcolor{red}{+\sigma^2_nI}]^{-1}(Y - \bm\mu) \\
\mathcal{K}_{Y|D} \triangleq \underbrace{\underbrace{\mathcal{K^{**}}}_\text{Prior} - \underbrace{\mathcal{K^*}^T[\mathcal{K} \textcolor{red}{+\sigma^2_nI}]^{-1}\mathcal{K^*}}_\text{Evidence}}_\text{Reduction in Variance}
\end{split}
\label{eq:noisyPred}
\end{align}
In Eq.~\eqref{eq:noisyPred} ${\bm\mu}^{*}$ represents the prior mean of the test inputs $X^*$ while $\bm{\mu}$ represents the mean of the noisy observations $Y$.

\subsection{Entropy}
The strength of GPs not only lies in the fact that they can be easily generalized to variety of spatial data by adjusting the nature of covariance kernel but also the fact that they give a measure of uncertainty. For model like Neural Networks, external methods need to be additionally deployed to measure the confidence of the model over its predictions but GPs already provide a measure of uncertainty \textit{i.e.,} \textit{Entropy}. Mathematically, it is given by:

\begin{align}
\begin{split}
\mathcal{H} &\triangleq \dfrac{1}{2} (2\pi\log e\sigma^2(\cdot)) \,,\\
 &= \underbrace{0.5\times(2\pi e)}_\text{Constant} + \underbrace{\log (\sigma(\cdot))}_\text{Std. Deviation}
\end{split}
\label{eq:Entropy}
\end{align}

For a proof sketch, the readers are referred to Section~\ref{app:entropy_GP}.

\subsection{Hyperparameters}
So far, we assumed that the distribution of the GP prior was \textit{a priori} given as given in Eq.~\eqref{eq:sq_exp_cov}. However, the prior distribution itself has free parameters called the \textit{hyperparameters (HPs)}. These include \textcolor{blue}{$\sigma_{sig}$} which represents the amplitude of the signal, \textcolor{red}{$\sigma_{n}^2$} which represents the noise variance. Besides these, there is an additional parameter called length-scale $(l)$ which represents the degree of smoothness of the covariance across the input space but has not been considered here. Thus, for the scope of this course, $\theta \triangleq \{\textcolor{blue}{\sigma_{sig}},\textcolor{red}{\sigma_{n}}\}$ and the noisy observations $Y\sim \mathcal{N}(0,\Sigma)$, where

\begin{align}
\Sigma \triangleq \textcolor{blue}{\sigma_{sig}^2} \underbrace{\mathcal{K}}_\text{ Eq.~\eqref{eq:sq_exp_cov}} \textcolor{red}{+ \sigma_n^2I}
\label{eq:sq_exp_cov_hps}
\end{align}

In order to accommodate the rate of decay of correlations across space, sometimes an additional hyper-parameter called the length scale $(l)$ is incorporated as follows:
\begin{align}
\Sigma(x,x') \triangleq \textcolor{blue}{\sigma_{sig}^2} \exp \left(\dfrac{-||x-x'||^2}{\textcolor{green}{l}}\right) \textcolor{red}{+ \sigma_n^2I}
\label{eq:sq_exp_cov_hps_all}
\end{align}

\subsection{Likelihood}
We started off with a 1-D prior and directly landed up with a posterior. However, as per Bayes rule, for carrying out Bayesian inference we have $Posterior\propto likelihood \times prior$. \begin{quote}
So, what happened to the likelihood?
\end{quote}
Likelihood, by definition, explains how likely it is to see the data points given the model that generated the data. First, let us consider a noise free case. Thus, we know that the data generating process is given by: $Y\sim \mathcal{N}(0,\textcolor{blue}{\sigma_{sig}}^2\mathcal{K}(\cdot,\cdot))$. The likelihood of this data generating MVN-pdf is then given by, 
\begin{equation}
\mathcal{L} = (2\pi\textcolor{blue}{\sigma_{sig}}^2)^{\frac{-n}{2}} |\mathcal{K}|^{\frac{-1}{2}}\exp\left\lbrace{\dfrac{-1}{2\textcolor{blue}{\sigma_{sig}}^2}Y^T\mathcal{K}^{-1}Y}\right\rbrace
\label{eq:likelihood}
\end{equation}
 where $n = \#(X)$ represents the cardinality \textit{i.e.,} the number of observed inputs. Taking the log of this, we get the log-likelihood which is given by:

\begin{equation}
\mathcal{LL} = \underbrace{-\dfrac{n}{2}\log(2\pi)-\dfrac{n}{2}\log(\textcolor{blue}{\sigma_{sig}}^2)}_\text{Const.}-\underbrace{\dfrac{1}{2}\log|\mathcal{K}|}_\text{Complexity}-\underbrace{\dfrac{1}{2\textcolor{blue}{\sigma_{sig}}^2}Y^T\mathcal{K}^{-1}Y}_\text{Data fit}
\label{eq:logLikelihood}
\end{equation}

In Eq.~\eqref{eq:logLikelihood}, the \textit{complexity} term penalizes the unnecessarily complicated models from being fit to the data \textit{i.e.,} Occam's razor principle while the \textit{data fit} terms penalizes the volume of the prior covariance. Now that the log-likelihood is defined, we can use this to learn $\theta$ by maximizing the log-likelihood using the type-II maximum likelihood estimation (MLE).

\section{Mathematical Tools}
In this section, we explain the key mathematical tools that are required to understand and efficiently implement the GP inference.

\subsection{Need for Jitter}
When the entries of the rows of a covariance matrix are very similar, matrix becomes ill-conditioned and inversion is also unstable (although we strictly advise not to attempt inversion). We first define the \textit{condition number (cond)} which can be used to evaluate how poorly conditioned is the matrix.

\begin{definition}[Condition Number (con)]
Consider the covariance matrix where the entries of the first $2$ rows are too similar as shown:

$ \begin{bmatrix}
1 & 0.9999 & 0 & 0 \\
0.9999 & 1 & 0 & 0 \\
0 & 0 & 1 & 0.1 \\
0 & 0 & 0.1 & 1 
\end{bmatrix}$
Then, $con = \dfrac{\lambda_{max}}{\lambda_{min}}$ where $\lambda$ represents the eigen values. In this case $con = 19999$ and higher the value, the more ill-conditioned the matrix becomes.
\end{definition}

Now the conditioning problem can be addressed by adding a small positive quantity to the diagonal entries as shown:$ \begin{bmatrix}
1.01 & 0.9999 & 0 & 0 \\
0.9999 & 1.01 & 0 & 0 \\
0 & 0 & 1.01 & 0.1 \\
0 & 0 & 0.1 & 1.01 
\end{bmatrix}$. The revised $con = 199$ since the entries are now sufficiently dissimilar. 

\textbf{N.B.:} Jitter essentially is adding noise to the data and hence adding unnecessarily large noise to data can dilute the informativeness of the data. Thus, the jitter must always be kept sufficiently small to avoid numerical instabilities whilst retaining the information to be processed.

\subsection{Avoiding Kernel Inversion}

In Eq.~\eqref{eq:posterior_cov}, we can see that the kernel $K$ needs to be inverted. Actually, inverting the kernel incurs the computational complexity of $\mathcal{O}(n^3)$ for a kernel of size $n\times n$. This grows exponentially as the size of kernel increases and this can be easily avoided by using Cholesky decomposition \cite{press2007numerical} instead which has the computational complexity of $\mathcal{O}(n^3)$ but inverting the Cholesky factor only incurs $\mathcal{O}(n^2)$.

\begin{exmp}[Cholesky Decomposition]
In this example, we will use Cholesky decomposition to solve a system of equations as opposed to direct matrix inversion which is computationally costlier as the size of matrix grows. Consider the following system of linear equations:
\begin{align}
\begin{split}
x_1 - x_2 + 2x_3 &= 17 \,,\\
-x_1 +5x_2 -4x_3 &= 31 \,,\\
2x_1 -4x_2 + 6x_3 &= -5 
\end{split}
\end{align}
In the matrix-vector notation, this can be written down as:
\begin{equation}
\label{eq:mat_vec}
\underbrace{\left[
  \begin{array}{ccc}
    1 & -1 & 2 \\
    -1 & 5 & -4 \\
    2 & -4 & 6 
  \end{array}
\right]}_\text{A}
\times
\underbrace{\begin{bmatrix}
           x_{1} \\
           x_{2} \\
           x_{3}
         \end{bmatrix}}_{\vec{\mathbf{x}}} = \underbrace{\begin{bmatrix}
           17 \\
           31 \\
           -5
         \end{bmatrix}}_{\vec{\mathbf{b}}}
\end{equation}

Now, a shorthand representation would be $A\mathbf{x}=\mathbf{b}$ where $A$ represents the coefficient matrix, $\mathbf{x}$ represents the vector of variables and $\mathbf{b}$ represents the vector of constants as also marked in Eq.~\eqref{eq:mat_vec}. Notice that the matrix $A$ is symmetric positive definite and hence the Cholesky decomposition can be utilized. If this was not the case, then a more generic variant called LU decomposition can be used herewith.

Mathematically,
\begin{equation}
A\mathbf{x}=\mathbf{b} \equiv LU\mathbf{x}=\mathbf{b}
\end{equation}

Here $LU$ refers to the L and U factor matrices of A. For this example,
\begin{equation}
L = \left[
  \begin{array}{ccc}
    1 & 0 & 0 \\
    -1 & 2 & 0 \\
    2 & -1 & 1 
  \end{array}
\right]
\end{equation}

and \begin{equation}
U = \left[
  \begin{array}{ccc}
    1 & -1 & 2 \\
    0 & 2 & -1 \\
    0 & 0 & 1 
  \end{array}
\right]
\end{equation}

Then, the system of equations can be written as:
\begin{equation}
\underbrace{\underbrace{\left[
  \begin{array}{ccc}
   1 & 0 & 0 \\
    -1 & 2 & 0 \\
    2 & -1 & 1  
  \end{array}
\right]}_\text{L}
\underbrace{\left[
  \begin{array}{ccc}
    1 & -1 & 2 \\
    0 & 2 & -1 \\
    0 & 0 & 1
  \end{array}
\right]}_\text{U}}_\text{A}
\underbrace{\begin{bmatrix}
           x_{1} \\
           x_{2} \\
           x_{3}
         \end{bmatrix}}_{\vec{\mathbf{x}}} = \underbrace{\begin{bmatrix}
           17 \\
           31 \\
           -5
         \end{bmatrix}}_{\vec{\mathbf{b}}}
\end{equation}

Now, to solve the original system of equations, we first solve the intermediate step of $LU\vec{\mathbf{x}}=\mathbf{b}$. For this, let $U\vec{\mathbf{x}}=\vec{\mathbf{y}}$ and solve $L\vec{\mathbf{y}}=\vec{\mathbf{b}}$. Once, the solution $\vec{\mathbf{y}}$ is obtained, substitute that back to get the values of $\vec{\mathbf{x}}$. For this example, we have

\begin{equation}
\left[
  \begin{array}{ccc}
    1 & 0 & 0 \\
    -1 & 2 & 0 \\
    2 & -1 & 1  
  \end{array}
\right]
\begin{bmatrix}
           y_{1} \\
           y_{2} \\
           y_{3}
         \end{bmatrix} = \begin{bmatrix}
           17 \\
           31 \\
           -5
         \end{bmatrix}
\end{equation}

Solving which gives us
\begin{equation}
\begin{bmatrix}
           y_{1} \\
           y_{2} \\
           y_{3}
         \end{bmatrix} = \begin{bmatrix}
           17 \\
           24 \\
           -15
         \end{bmatrix}
\end{equation}

This implies that, 
\begin{equation}
\left[
  \begin{array}{ccc}
    1 & -1 & 2 \\
    0 & 2 & -1 \\
    0 & 0 & 1 
  \end{array}
\right]
\begin{bmatrix}
           x_{1} \\
           x_{2} \\
           x_{3}
         \end{bmatrix} = \begin{bmatrix}
           17 \\
           24 \\
           -15
         \end{bmatrix}
\end{equation}

Solving this, finally returns the original unknown as:
\begin{equation}
\begin{bmatrix}
           x_{1} \\
           x_{2} \\
           x_{3}
         \end{bmatrix} = \begin{bmatrix}
           51.5 \\
           4.5 \\
           -15.0
         \end{bmatrix}
\end{equation}
\end{exmp}
Owing to the sparsity of the upper and lower triangular factors, matrix manipulations are much more memory efficient. Also, note that $U = L^T$ and thus, computationally only one factor needs to be computed and stored in memory. For the other factor, the previous one simply needs to be transposed. Like this, the need for actual matrix inversion can be by-passed in a computationally efficient and stable way.

\subsection{Implementing Gradients}
We need to differentiate the kernel with respect to each of its hyper-parameters to deduce the optimal values using the \textit{Maximum Likelihood Estimation}. The derivations are given in Eq.~\eqref{eq:dk_dsig}-Eq.~\eqref{eq:dk_dnoise} in the Appendix.
\section{Hands-on Exercises with \texorpdfstring{$1D$}{1D} GPs}
In this section, we present the detailed discussions of the hands-on programming exercises. For the ease of understanding, we begin with the simplest case of 1 dimensional analysis. Thus, our inputs $x \in \mathcal{R}^1$ and targets $y \in \mathcal{R}^1$. 

\subsection{Task 1: Generating the inputs and targets}
\begin{lstlisting}[language=Python, caption=Generate input data]
# Import all necessary modules here
import numpy as np
import math
import matplotlib.pyplot as pl
from scipy.optimize import fmin_l_bfgs_b as bfgs
from scipy.optimize import minimize 

# Number of data samples
numObs = 8 # Number of Observations
numTest = 100 # Number of testing inputs

# Jitter Quantities
eps = 1.49e-08 ## really small jitter for numerical stability


# Noiseless training data
Xtrain = np.linspace(start=0, stop=2*math.pi, num=numObs)[:, np.newaxis] # Training Inputs
ytrain = np.sin(Xtrain) # Training Targets

fig, ax = pl.subplots()
pl.plot(Xtrain, ytrain)
pl.title('Training Signal')
pl.show()

# Test data
Xtest = np.linspace(start=-0.5, stop=2*math.pi+0.5, num=numTest)[:, np.newaxis]
\end{lstlisting}

\subsection{Task 2: Now implement a function that returns the sqaured eucledian distance amongst all possible input pairs. The output should be a square matrix.}
\label{task:distmat}
\begin{lstlisting}[language=Python, caption=Squared Distance Function]
# Define the kernel function
def Sq_Euclid_DistMat(X1,X2):
    '''
    L2-norm applicable for both vectors and matrices (useful for high dimension input features.)
    Parameter Description:
    X1,X2: When X1==X2, then calculate autocorrelation. Otherwise cross-correlations.
    DistMat: Pairwise Squared Distance Matrix of size nXn
    '''
    if X1.shape[1] == 1:  # vectors
        n = X1.shape[0]
        m = X2.shape[0]
        r1 = X1.reshape(n, 1) * np.ones([1, m])
        r2 = X2.reshape(1, m) * np.ones([n, 1])
        sed = ((r1 - r2) ** 2)
    elif X1.shape[1] == 2:  # matrices for 2D feature space.
        n = X1.shape[0]
        m = X2.shape[0]
        r1x = X1[:, 0].reshape(n, 1) * np.ones([1, m])
        r1y = X1[:, 1].reshape(n, 1) * np.ones([1, m])
        r2x = X2[:, 0].reshape(1, m) * np.ones([n, 1])
        r2y = X2[:, 1].reshape(1, m) * np.ones([n, 1])
        sed = ((r1x - r2x) ** 2 + (r1y - r2y) ** 2)
    else:
        print ("too many dimensions in X matrices", X1.shape)
        return None

    return sed
\end{lstlisting}

\subsection{Task 3: Now make draws from a multivariate Gaussian pdf with Mean as 0 and covariance given by the RBF Kernel from above. Start off with 1 draw and then draw 5 priors.}

\begin{lstlisting}[language=Python, caption=Squared Exponential Kernel]
np.random.seed(1) # set seed for consistency

DXtest = Sq_Euclid_DistMat(Xtest,Xtest)
K_ss = np.exp(-DXtest) # Auto-correlation between test points K(X*,X*)

# Get the lower cholesky factor of the covariance matrix
L = np.linalg.cholesky(K_ss+eps*np.eye(numTest))

## Make 1 draw
f_prior1 = np.dot(L, np.random.normal(size=(numTest,1)))

# Plotting the drawn prior.
fig, ax = pl.subplots()
pl.plot(Xtest, f_prior1)
pl.title('One sample from the GP prior')
pl.show()
fig.savefig('../Figures/1Draw.png', bbox_inches='tight')

## Now make 5 draws
f_prior5 = np.dot(L, np.random.normal(size=(numTest,5)))

# Plotting all 5 priors.
fig, ax = pl.subplots()
pl.plot(Xtest, f_prior5)
pl.title('Five samples from the GP prior')
pl.show()
\end{lstlisting}

\begin{figure}[!htbp]
\centering
\includegraphics[scale=0.8]{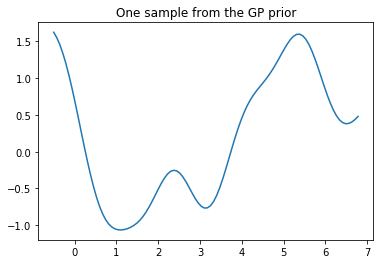}
\caption{Drawing $1$ sample from the GP Prior.}
\end{figure}

\begin{figure}[!htbp]
\centering
\includegraphics[scale=0.8]{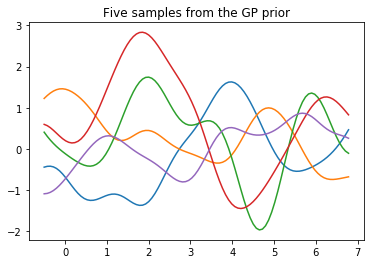}
\caption{Drawing $5$ samples from the GP Prior.}
\end{figure}

\subsection{Task 4: Now, let us make a posterior for 100 samples that were not used previously for training the GP.}

\begin{lstlisting}[language=Python, caption=Drawing priors from GPs]
# Generate all necessary kernels for predictions
D = Sq_Euclid_DistMat(Xtrain,Xtrain) # Squared Eucledian Distance
K = np.exp(-D + np.diag(eps*np.ones(numObs))) # K(X,X)
fig, ax = pl.subplots()
pl.imshow(K)
ax.set_ylim(ax.get_ylim()[::-1]) 
pl.colorbar()
pl.title(u'$K(X,X)$')


fig, ax = pl.subplots()
pl.imshow(K_ss)
ax.set_ylim(ax.get_ylim()[::-1])
pl.colorbar()
pl.title(u'$K(X^*,X^*)$')


# Apply the kernel function to our training points
L = np.linalg.cholesky(K) # lower cholesky factor 

# Compute the mean at our test points.
DX = Sq_Euclid_DistMat(Xtrain, Xtest)
K_s = np.exp(-DX) # Cross Correlation between test and train inputs
fig, ax = pl.subplots()
pl.imshow(K_s)
ax.set_ylim(ax.get_ylim()[::-1]) 
pl.colorbar()
pl.title(u'$K(X,X*)$')

Lk = np.linalg.solve(L, K_s)
mu = np.dot(Lk.T, np.linalg.solve(L, ytrain)).reshape((numTest,))

# Compute the standard deviation to find the upper and lower Quantiles
s2 = np.diag(K_ss) - np.sum(Lk**2, axis=0) # variance
stdv = np.sqrt(s2) # std deviation

# Draw samples from the posterior at the test points.
L = np.linalg.cholesky(K_ss+eps*np.eye(numTest) - np.dot(Lk.T, Lk)) # add small jitter to keep the kernel psd
f_post = mu.reshape(-1,1) + np.dot(L, np.random.normal(size=(numTest,3)))

# Generate Plots
fig, ax = pl.subplots()
pl.plot(Xtrain, ytrain, 'bs', ms=8,label=u'$f(x) = \sin(x)$') # Original Data
pl.plot(Xtest, f_post) # Posterior Samples
pl.gca().fill_between(Xtest.flat, mu-2*stdv, mu+2*stdv, color="#dddddd",label='95% CI') # 95% CI
pl.plot(Xtest, mu, 'r--', lw=2,label=u'$\mu$') # Posterior Mean
ax.legend(loc='best', fancybox=True, framealpha=0.5)
pl.title('Three samples from the GP posterior')
pl.show()
\end{lstlisting}

\begin{figure}[!htbp]
\label{fig:1D_posterior}
\centering
\includegraphics[scale=0.8]{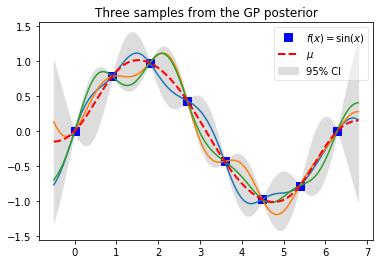}
\caption{1D Posterior.}
\end{figure}

\subsection{Task 5: Now the confidence bounds look acceptable and we are satisfied with the data fit of the posterior GP. However, what is likely to happen should we simply decide to scale the data by a constant factor? For this task, replace $f(x) = \sin (x)$ with $f(x) = 5 \times \sin (x)$ to see how the GP behaves.} 
\begin{lstlisting}[language=Python, caption=Scaling the Input data]
  yScaled = 5*np.sin(Xtrain)  # Scaled Targets

# Apply the kernel function to our training points
D = Sq_Euclid_DistMat(Xtrain,Xtrain) # Squared Eucledian Distance
K = np.exp(-D + np.diag(eps*np.ones(numObs))) # K(X,X)
L = np.linalg.cholesky(K) # lower cholesky factor


muScaled = np.dot(Lk.T, np.linalg.solve(L, yScaled)).reshape((numTest,))

# Compute the standard deviation to find the upper and lower Quantiles
s2 = np.diag(K_ss) - np.sum(Lk**2, axis=0)
stdv = np.sqrt(s2)

# Draw samples from the posterior at the test points.
L = np.linalg.cholesky(K_ss + 1e-6*np.eye(numTest) - np.dot(Lk.T, Lk))
f_post = muScaled.reshape(-1,1) + np.dot(L, np.random.normal(size=(numTest,3)))

# Generate Plots
fig, ax = pl.subplots()
pl.plot(Xtrain, yScaled, 'bs', ms=8,label=u'$f(x) = 5*\sin(x)$') # Original Data
pl.plot(Xtest, f_post) # Posterior Samples
pl.gca().fill_between(Xtest.flat, muScaled-2*stdv, muScaled+2*stdv, color="#dddddd",label='95% CI') # 95% CI
pl.plot(Xtest, muScaled, 'r--', lw=2,label=u'$\mu$') # Posterior Mean
ax.legend(loc='best', fancybox=True, framealpha=0.5)
pl.title('Fitting original GP to scaled data')
pl.show()
\end{lstlisting}

\begin{figure}[!htbp]
\label{fig:GP_scaled}
\centering
\includegraphics[scale=0.8]{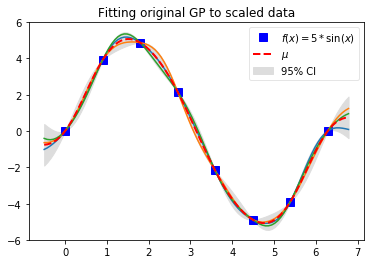}
\caption{Fitting the original GP as it is to the scaled data.}
\end{figure}

\subsection{Task 6: In real world, the data available is usually noisy. What happens if we try to fit a GP to a noisy data? Add a white noise to the previously generated data and fit the original GP to the data.}
\begin{lstlisting}[language=Python, caption=Fitting GP to noisy observations]
# Generate noisy targets
noise_scale = 0.4
noise_mean = 0 # zero mean gaussian noise
noise_var = 1 # true variance of noise being added
noise = noise_scale * np.random.normal(noise_mean,noise_var,numObs)[:, np.newaxis]
Y_noisy = ytrain + noise # Noisy targets

# Fitting the original GP here.
D = Sq_Euclid_DistMat(Xtrain,Xtrain) # Squared Eucledian Distance
K = np.exp(-D + np.diag(eps*np.ones(numObs))) # K(X,X)
L = np.linalg.cholesky(K) # lower cholesky factor

muNoisy = np.dot(Lk.T, np.linalg.solve(L, Y_noisy)).reshape((numTest,))

# Compute the standard deviation to find the upper and lower Quantiles
s2 = np.diag(K_ss) - np.sum(Lk**2, axis=0)
stdv = np.sqrt(s2)

# Draw samples from the posterior at the test points.
L = np.linalg.cholesky(K_ss + 1e-6*np.eye(numTest) - np.dot(Lk.T, Lk))
f_post = muNoisy.reshape(-1,1) + np.dot(L, np.random.normal(size=(numTest,3)))

# Generate Plots
fig, ax = pl.subplots()
pl.plot(Xtrain, Y_noisy, 'bs', ms=8,label=u'$f(x) = \sin(x) + \epsilon$') # Original Data
pl.plot(Xtest, f_post) # Posterior Samples
pl.gca().fill_between(Xtest.flat, muNoisy-2*stdv, muNoisy+2*stdv, color="#dddddd",label='95% CI') # 95% CI
pl.plot(Xtest, muNoisy, 'r--', lw=2,label=u'$\mu$') # Posterior Mean
ax.legend(loc='best', fancybox=True, framealpha=0.5)
pl.title('Fitting original GP to noisy data')
pl.show()
\end{lstlisting}

\begin{figure}[!htbp]
\centering
\includegraphics[scale=0.8]{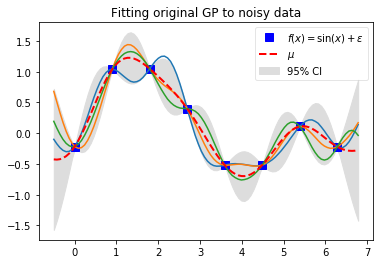}
\caption{Fitting original GP to noisy targets.}
\end{figure}

\subsection{Task 7: We have already introduced amplitude and noise parameters before or more precisely \textit{hyper-parameters}. Now, using the Maximum Likelihood Estimation (MLE), try to infer the data scale $\hat{\tau} = 2\hat{\sigma}_{sig}$ that best fit the data. As a sanity check, remember that the original data was scaled by a factor of $5$.}
\begin{lstlisting}[language=Python, caption=Maximum Likelihood Estimation (MLE) for $\sigma_{sig}$]
c = np.linalg.inv(np.linalg.cholesky(K))
Ci = np.dot(c.T,c) 

C_ss=K_ss

sigma_sigSq = np.dot(np.dot((yScaled).T,Ci),yScaled) / numObs
EstDataScale = 2*np.sqrt(sigma_sigSq)
print("Data Scale Estimated as: ",EstDataScale[0][0])

# Now fit the scaled GP to scaled data
# Apply the kernel function to our training points
D = Sq_Euclid_DistMat(Xtrain,Xtrain) # Squared Eucledian Distance
K = np.exp(-D + np.diag(eps*np.ones(numObs))) # K(X,X)
L = np.linalg.cholesky(K) # lower cholesky factor


muScaled = np.dot(Lk.T, np.linalg.solve(L, yScaled)).reshape((numTest,))

# Compute the standard deviation to find the upper and lower Quantiles
s2 = sigma_sigSq[0]*(np.diag(K_ss) - np.sum(Lk**2, axis=0))
stdv = np.sqrt(s2)

# Draw samples from the posterior at the test points.
L = np.linalg.cholesky(K_ss + 1e-6*np.eye(numTest) - np.dot(Lk.T, Lk))
f_post = muScaled.reshape(-1,1) + np.dot(L, np.random.normal(size=(numTest,3)))

# Generate Plots
fig, ax = pl.subplots()
pl.plot(Xtrain, yScaled, 'bs', ms=8,label=u'$f(x) = 5*\sin(x)$') # Original Data
pl.plot(Xtest, f_post) # Posterior Samples
pl.gca().fill_between(Xtest.flat, muScaled-2*stdv, muScaled+2*stdv, color="#dddddd",label='95% CI') # 95% CI
pl.plot(Xtest, muScaled, 'r--', lw=2,label=u'$\mu$') # Posterior Mean
ax.legend(loc='best', fancybox=True, framealpha=0.5)
pl.title('Fitting scaled GP to scaled data')
pl.show()
\end{lstlisting}

\begin{figure}[!htbp]
\centering
\includegraphics[scale=0.8]{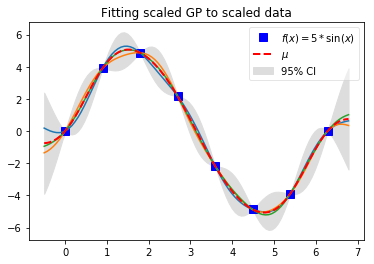}
\caption{Fitting scaled GP to scaled noise-free targets.}
\end{figure}

\subsection{Task 8: Now, using the Maximum Likelihood Estimation (MLE), try to infer the noise variance $\hat{\sigma}_{n}$. Sometimes in literature these may also be referred to as nuggets which we see as the grey blobs in our figures here that represent confidence bounds.}
\begin{lstlisting}[language=Python, caption=Maximum Likelihood Estimation (MLE) for $\sigma_{n}$]
def nll_n(sig_n,DistMat,Y):
    '''
    Function to obtain the negative log likelihood for the noise HP
    sig_n: noise std. dev
    DistMat: Pairwise Eucledian Distance Matrix
    Y: List of targets/ Observations
    
    :return: negative log likelihood -ll
    '''
    K = np.exp(-DistMat) + np.diag(sig_n**2*np.ones(numObs))
    c = np.linalg.inv(np.linalg.cholesky(K))
    Ki = np.dot(c.T,c) 
    (sign, logdetK) = np.linalg.slogdet(K)
    ll = -numObs/2 * np.log(np.dot(Y.T,(np.dot(Ki,Y)))) - 1/2 * logdetK
    return -ll

def gnll_n(sig_n,DistMat,Y):
    '''
    Function to obtain gradient of negative log-likelihood for the noise HP
    sig_n: noise std. dev
    DistMat: Pairwise Eucledian Distance Matrix
    Y: List of targets/ Observations
    
    :return: negative gradients of negative log likelihood -dll
    '''
    K = np.exp(-DistMat) + np.diag(sig_n**2*np.ones(numObs))
    c = np.linalg.inv(np.linalg.cholesky(K))
    Ki = np.dot(c.T,c)
    KiY = np.dot(Ki,Y)
    dll = (numObs)/2 * np.dot(KiY.T,KiY)/(np.dot(Y.T,KiY)) - 1/2*(np.sum(np.diag(Ki)))
    return -dll
   
YScaled_noisy = yScaled + noise # Noisy scaled targets
noise_bounds = [(eps, np.var(YScaled_noisy))]   # lower and upper-bounds for noise variance 
initial_guess = 0.1*np.var(YScaled_noisy)
DistMat = Sq_Euclid_DistMat(Xtrain,Xtrain)
res = minimize(nll_n, initial_guess,args=(DistMat,YScaled_noisy), method="L-BFGS-B",\
               jac=gnll_n, bounds = noise_bounds,options={'maxiter':1000, 'gtol': 1e-6, 'disp': True})

sig_n = np.sqrt(res.x[0]) # Noise std. dev
print("Optimal Noise Hyper-parameter: ",sig_n) # best theta
print("Optimal Negative Log-Likelihood: ",res.fun[0][0]) # log marginal likelihood.
print("Convergence Status: ",res.message) 
print("No. of Evaluations: ", res.nfev)

# Plug back to find tau_hat*2
K = np.exp(-DistMat) + np.diag(sig_n**2*np.ones(numObs))
c = np.linalg.inv(np.linalg.cholesky(K))
Ki = np.dot(c.T,c)
sigma_sigSq = np.dot(np.dot((YScaled_noisy).T,Ci),YScaled_noisy) / numObs
EstDataScale = 2*np.sqrt(sigma_sigSq)
print("Data Scale Estimated as: ",EstDataScale[0][0])
\end{lstlisting}

\subsection{Task 9: In order to estimate the smoothness of the variation in correlations across space, we usually utilize a parameter called spatial length scale $l$.  Now, using the Maximum Likelihood Estimation (MLE), try to infer the spatial length scale $l$ and noise variance $\sigma_{n}$.}
\begin{lstlisting}[language=Python, caption=Maximum Likelihood Estimation (MLE) for $l$ and $\sigma_{n}$]
def nll(params,DistMat,Y):
    '''
    Function to obtain the negative log likelihood for the length scale and noise
    params: length scale and noise std_dev as a list
    DistMat: Pairwise Eucledian Distance Matrix
    Y: List of targets/ Observations
    
    :return: negative log likelihood -ll
    '''
    length_scale = params[0] ## Account for length scale here
    sig_n = params[1] ## Noise std. deviation
    K = np.exp(-DistMat/length_scale) + np.diag(sig_n**2*np.ones(numObs)) ## Note the change here
    c = np.linalg.inv(np.linalg.cholesky(K+eps*(np.eye(numObs))))
    Ki = np.dot(c.T,c) 
    (sign, logdetK) = np.linalg.slogdet(K)
    ll = -numObs/2 * np.log(np.dot(Y.T,(np.dot(Ki,Y)))) - 1/2 * logdetK
    return -ll

def gnll(params,DistMat,Y):
    '''
    Function to obtain gradient of negative log-likelihood
    sig_n: noise std. dev
    DistMat: Pairwise Eucledian Distance Matrix
    Y: List of targets/ Observations
    
    :return: negative gradients of negative log likelihood -dll
    '''
    length_scale = params[0] ## Account for length scale here
    sig_n = params[1]
    K = np.exp(-DistMat/length_scale) + np.diag(sig_n**2*np.ones(numObs)) ## Note the change here
    c = np.linalg.inv(np.linalg.cholesky(K+eps*(np.eye(numObs))))
    Ki = np.dot(c.T,c)
    KiY = np.dot(Ki,Y)
    dotK = np.dot(K,DistMat)/(length_scale**2)
    
    # Compute derivatives for both components separately
    dll_ls = (numObs)/2 * np.dot(KiY.T,np.dot(dotK,KiY))/(np.dot(Y.T,KiY))\
            - (1/2)*(np.sum(np.diag(np.dot(Ki,dotK)))) # for length scale
    dll_n = (numObs)/2 * np.dot(KiY.T,KiY)/(np.dot(Y.T,KiY)) - 1/2*(np.sum(np.diag(Ki))) # for noise
    return(np.concatenate((dll_ls, dll_n), axis=0))
    

initial_guess = [(0.1,0.1*np.var(YScaled_noisy))]
HP_bounds = ((eps, 10),(eps, np.var(YScaled_noisy)))


DistMat = Sq_Euclid_DistMat(Xtrain,Xtrain)
res = minimize(nll, initial_guess,args=(DistMat,YScaled_noisy), method="L-BFGS-B",\
               jac=gnll,bounds = HP_bounds,options={'maxiter':1000, 'gtol': 1e-6, 'disp': True})

print("Optimal length scale: ",res.x[0]) 
print("Optimal Noise Variance: ",res.x[1])
print("Optimal Negative Log-Likelihood: ",res.fun[0][0]) # log marginal likelihood.
print("Convergence Status: ",res.message) 
print("No. of Evaluations: ", res.nfev)
\end{lstlisting}
\section{Discussion}
Here, we discuss the results obtained above after performing the programming exercises. We obtained posterior (predictive distribution) for the dummy $1D$ data and this is shown in Fig.~\ref{fig:1D_posterior}. Here, the predictor is interpolating over the data and the ``football'' like shapes represent the error bars. There is almost no error in prediction at the observations (blue squares) but the errors get bigger as the predictor tries to make a prediction farther away from the observations. This can be attributed to the fact that the correlation decays as the spatial separation increases between the observations and the test point. Thus, if the prediction is to be made at test points that are sufficiently far away, the predictor gets more uncertain about the predictions. However, the predictive mean is always mean-reverting (converges back to zero). In Fig.~\ref{fig:GP_scaled}, we tried to fit the original GP with an amplitude of $1$ to a data with a larger amplitude. Although, the posterior seems to follow the trend appropriately but the truth is revealed when the confidence quantiles are analyzed. Despite being able to fit the posterior mean perfectly, the choice of the wrong prior took its toll. The GP here, is underestimating its variance (over confident) which is not good for practical applications. This, was easily rectified by either scaling the GP by eye-balling or performing MLE.

\section{Conclusions}
The aim of this document was to give a crash-course to its users pertaining to the domain of Gaussian Processes (GPs). We sincerely hope that by the end of this crash course, the users will be able to understand how the GPs work instead of simply deploying them as black boxes. All the code provided herewith is available in an interactive environment on our \href{https://sites.google.com/view/exposure-to-gp/}{website}. The users are encouraged to try and tinker with the code to enhance their understanding and customize the implementation to their liking.

\section{Future Works}
In the further courses in this series, we will focus on extending the input dimension to $2D$ data with $1D$ targets and provide with hands-on exercises yet again. We will also improve upon our existing course based on the user \href{https://sites.google.com/view/exposure-to-gp/feedback}{feedback} we accrue with the passage of time. If you wish to \href{https://sites.google.com/view/exposure-to-gp/feedback/contribute}{contribute}, please fill out the contribution request form on our website to let us know what and how you would like to contribute.
\section{Acknowledgement}
The author would like to thank Professor Robert B. Gramacy of the Department of Statistics, Virginia Tech for the discussions.

\cleardoublepage
\section{Appendix}
\appendix
\renewcommand*{\thesection}{\Alph{section}}
\subsection{Inference using stationary GPs with RBF Kernels}
\label{app:GP_Inf}
\begin{lemma}[Inverse of Partitioned Matrix]
If $A$ is non-singular $n\times n$ matrix partitioned as $A = \begin{bmatrix}
       A_{11} & A_{12} \\[0.3em]
       A_{21} & A_{22} \\[0.3em]
    \end{bmatrix}
$. Then, the inverse of this partitioned matrix is given by:
\begin{equation}
A^{-1} = \begin{bmatrix}
       (A_{11}-A_{12}A_{22}^{-1}A_{21})^{-1} & -(A_{11}-A_{12}A_{22}^{-1}A_{21})^{-1}A_{12}A_{22}^{-1} \\[0.3em]
       -(A_{22}-A_{21}A_{11}^{-1}A_{12})^{-1}A_{21}A_{11}^{-1} & (A_{22}-A_{21}A_{11}^{-1}A_{12})^{-1} \\[0.3em]
     \end{bmatrix}
\end{equation}
\label{lemma:InvpartitionMat}
\end{lemma}
\begin{proof}
For proof sketch, please refer to \cite{InvPartitionMat}.
\end{proof}

\begin{theorem}[Posterior over Exponential Kernels]
\label{app:Posterior_Proof}
Given a training dataset $D = [X,Y]$ where $X$ represents the inputs in $\mathcal{R}^n$ and $Y\in \mathcal{R}^m$ $(m \ll n)$ represent the corresponding targets, a GP model can predict the measurements for any previously unobserved set of inputs $(X^*)$ using the predictive distribution $p(\mathbf{y}^*|X^*,D)\sim \mathcal{N}(\bm{\mu}_{Y^*|D},\mathcal{K}_{Y^*|D})$. Here,
\begin{align}
\begin{split}
\bm{\mu}_{Y^*|D} &= \cancelto{0}{\bm\mu}^{*} + \mathcal{K^*}^T\mathcal{K}^{-1}(Y - \bm\mu) \\
\mathcal{K}_{Y^*|D} &=  \mathcal{K^{**}} - \mathcal{K^*}^T\mathcal{K}^{-1}\mathcal{K^*}
\end{split}
\end{align}
\end{theorem}

\begin{proof}
Our noisy observations $\{(\mathbf{x}_i,y_i)\}_{i=1}^N$ for $\mathbf{x}\in \mathbb{R}^D$ and $y\in \mathbb{R}^1$ can be represented using some latent function $f$ as:
\begin{equation}
y_i = f({\mathbf{x}_i})+ \epsilon_i
\end{equation}
where $f\sim GP(\mu(\cdot),k_f(\cdot,\cdot))$\footnote{Here $k_f$ represents the kernel such that $\textbf{cov}(f(x),f(x')) = k_f(x,x')$} and $\epsilon_i \sim \mathcal{N}(0,\sigma^2)$. Consider a set of observed inputs $\mathbf{x}\in X$ and unobserved inputs $\mathbf{x}^*\in X^*$. Since the sum of independent Gaussian random variables is also Gaussian, we have:
\begin{align}
\begin{split}
\begin{bmatrix}
\mathbf{y}\\
\mathbf{y^*}
\end{bmatrix}
&= \begin{bmatrix}
f(\mathbf{x})\\
f(\mathbf{x^*})
\end{bmatrix} 
+
\begin{bmatrix}
\bm\epsilon\\
\bm\epsilon^*
\end{bmatrix} \\
&= \mathcal{N}\left(\begin{bmatrix}
\bm\mu\\
\bm\mu^{*}
\end{bmatrix} 
,
\begin{bmatrix}
\mathcal{K}(X,X) & \mathcal{K}(X,X^*) \\[0.3em]
\mathcal{K}(X,X^*)^T & \mathcal{K}(X^*,X^*) \\[0.3em]
\end{bmatrix}\right) + 
\mathcal{N}\left(\begin{bmatrix}
\vec{0}\\
\vec{0}
\end{bmatrix} 
,
\begin{bmatrix}
\sigma^2I & \vec{0} \\[0.3em]
\vec{0} & \vec{0} \\[0.3em]
\end{bmatrix}\right) \\
&=\mathcal{N}\left(\begin{bmatrix}
\bm\mu\\
\bm\mu^{*}
\end{bmatrix} 
,
\begin{bmatrix}
\mathcal{K}(X,X)+\sigma^2I & \mathcal{K}(X,X^*) \\[0.3em]
\mathcal{K}(X,X^*)^T & \mathcal{K}(X^*,X^*)+\sigma^2I \\[0.3em]
\end{bmatrix}\right) 
\end{split}
\end{align}

Let $A = \begin{bmatrix}
\mathcal{K} + \textcolor{red}{\sigma_n^2I} & \mathcal{K}(X,X^*) \\[0.3em]
\mathcal{K}(X,X^*)^T & \mathcal{K}(X^*,X^*) \\[0.3em]
\end{bmatrix} $ represent the partitioned matrix used above and $V=A^{-1}$ represent the inverse of such matrix. Then, from Lemma~\ref{lemma:InvpartitionMat}, we can obtain the inverse of this partitioned matrix such that:
\begin{align}
V=A^{-1} &= \begin{bmatrix}
V & V^{*} \\[0.3em]
{V^{*}}^T & V^{**}\\[0.3em]
\end{bmatrix}
\end{align}

which yields the following quantities,
\begin{align}
\begin{split}
V &=  ([\mathcal{K} + \textcolor{red}{\sigma_n^2I}]-\mathcal{K}^{*}{\mathcal{K}^{**}}^{-1}{\mathcal{K}^{*}}^T)^{-1}\,,\\
V^{*} &= -([\mathcal{K} + \textcolor{red}{\sigma_n^2I}]-\mathcal{K}^{*}{\mathcal{K}^{**}}^{-1}{\mathcal{K}^{*}}^T)^{-1}\mathcal{K}^{*}{\mathcal{K}^{**}}^{-1}\,,\\
{V^{*}}^T &= -(\mathcal{K}^{**}-{\mathcal{K}^{*}}^T[\mathcal{K} + \textcolor{red}{\sigma_n^2I}]^{-1}\mathcal{K}^{*})^{-1}{\mathcal{K}^{*}}^T\mathcal{K}^{-1}\,,\\
V^{**} &= (\mathcal{K}^{**}-{\mathcal{K}^{*}}^T[\mathcal{K} + \textcolor{red}{\sigma_n^2I}]^{-1}\mathcal{K}^{*})^{-1}\,.
\end{split}
\label{eq:InvMat}
\end{align}

So, the posterior on $Y^*$ is now given by the conditional probability $p[\mathbf{y}^*\in Y^*|X^*,D]$ which can be expanded as:
\begin{align}
\begin{split}
&p[\mathbf{y}^*\in Y^*|X^*,D] = \\ &\frac{1}{\zeta_1}\cdot\left[\exp \left\lbrace-\frac{1}{2}\left(\begin{bmatrix}
           Y^* \\
           Y
         \end{bmatrix}-\begin{bmatrix}
           \bm\mu^{*} \\
           \bm\mu
         \end{bmatrix}\right)^T \begin{bmatrix}
V & V^{*} \\[0.3em]
{V^{*}}^T & V^{**}\\[0.3em]
\end{bmatrix}\left(\begin{bmatrix}
           Y^* \\
           Y
         \end{bmatrix}-\begin{bmatrix}
           \bm\mu^{*} \\
           \bm\mu
         \end{bmatrix}\right)\right\rbrace \right]
\end{split}
\end{align}
where, $\zeta_1$ is the normalization constant independent of $Y^*$.

We will now expand the expression within the $\exp\{\cdot\}$ in an attempt to simplify it. Thus far, we have:
\begin{align}
\begin{split}
&\left(\begin{bmatrix}
           Y^* \\
           Y
         \end{bmatrix}-\begin{bmatrix}
           \bm\mu^{*} \\
           \bm\mu
         \end{bmatrix}\right)^T \begin{bmatrix}
V & V^{*} \\[0.3em]
{V^{*}}^T & V^{**}\\[0.3em]
\end{bmatrix}\left(\begin{bmatrix}
           Y^* \\
           Y
         \end{bmatrix}-\begin{bmatrix}
           \bm\mu^{*} \\
           \bm\mu
         \end{bmatrix}\right)  \\
=&(Y^*-\bm\mu^{*})^TV^{**}(Y^*-\bm\mu^{*})+(Y^*-\bm\mu^{*})^T{V^{*}}^T(Y-\bm\mu)\\
& +(Y-\bm\mu)^TV^*(Y^*-\bm\mu^{*})+(Y-\bm\mu)^TV(Y-\bm\mu)^T       
\end{split}
\end{align}
We will retain only the terms dependent on $Y^*$ to get:
\begin{align}
p[\mathbf{y}^*\in Y^*|X^*,D] = \frac{1}{\zeta_2}\exp\left(-\frac{1}{2}\left[{Y^*}^TV^{**}Y^*-2{Y^*}^TV^{**}\bm\mu^{*}+2{Y^*}^T{V^{*}}^T(Y-\bm{\mu})\right] \right)
\end{align}

By completing the square, we can further simplify the expression to:
\begin{align}
p[\mathbf{y}^*\in Y^*|X^*,D] = \frac{1}{\zeta_3}\exp\left(-\frac{1}{2}(Y^*-\bm\mu_{Y^*|D})^TV^{**}(Y^*-\bm\mu_{Y^*|D})\right)
\end{align}
where $\bm\mu_{Y^*|D} = \bm\mu^{*}-{V^{**}}^{-1}{V^{*}}^{T}(Y-\bm\mu)$.

Thus , 
\begin{align}
p[\mathbf{y}^*\in Y^*|X^*,D]\sim&\mathcal{N}(\bm\mu_{Y^*|D},{V^{**}}^{-1})
\end{align}
Now, plugging Eq.~\eqref{eq:InvMat} above, we get:
\begin{align}
\begin{split}
\bm\mu_{Y^*|D} &= \bm\mu^*-{V^{**}}^{-1}{V^*}^T(Y-\bm\mu) \\
&= \bm\mu^*+ (\mathcal{K}^{**}-{\mathcal{K}^{*}}^T[\mathcal{K} + \textcolor{red}{\sigma_n^2I}]^{-1}\mathcal{K}^{*}) \\
&(\mathcal{K}^{**}-{\mathcal{K}^{*}}^T[\mathcal{K} + \textcolor{red}{\sigma_n^2I}]^{-1}\mathcal{K}^{*})^{-1} \\
&{\mathcal{K}^{*}}^T[\mathcal{K} + \textcolor{red}{\sigma_n^2I}]^{-1}(Y-\bm\mu) \\
&= \bm\mu^*+ {\mathcal{K}^{*}}^T[\mathcal{K} + \textcolor{red}{\sigma_n^2I}]^{-1}(Y-\bm\mu)
\end{split}
\end{align}
and
\begin{equation}
{V^{**}}^{-1} = \mathcal{K}^{**} - {\mathcal{K}^*}^T[\mathcal{K} + \textcolor{red}{\sigma_n^2I}]^{-1}\mathcal{K}^*
\end{equation}

\end{proof}

\begin{lemma}[MLE using RBF Kernel]
The likelihood of seeing a noisy observation $y = f(x) + \epsilon$ is defined as $p(Y|D,\theta)$ where $X$ represents the observed inputs and $\theta$ represents the hyper-parameters of the RBF kernel defined by Eq.~\eqref{eq:sq_exp_cov_hps}. Then, the log likelihood is given by:
\begin{equation}
\mathcal{LL} = \log(p(\mathbf{y}|X,\bm{\theta})) = \underbrace{-\frac{1}{2}\log|\mathcal{K}|}_\text{model complexity} \underbrace{-\frac{1}{2}\mathbf{y}^T \mathcal{K}^{-1}\mathbf{y}}_\text{model fit} \underbrace{-\frac{1}{2}\log(2\pi)}_\text{normalizer} 
\end{equation}
\label{lemma:lml}
The optimal hyper-parameters for the RBF kernel are those which maximize the marginal log-likelihood given in Eq.~\eqref{lemma:lml}. Thus, we define the partial derivatives of $\mathcal{L}$ with respect to the hyper-parameters ${\theta}$ as:
\begin{equation}
\begin{split}
\frac{\partial \mathcal{LL}}{\partial \theta_i} &= -\frac{1}{2}\frac{\partial \log|\mathcal{K}|}{\partial \theta_i} -\frac{1}{2}\frac{\partial \mathbf{y}^T \mathcal{K}^{-1}\mathbf{y}}{\partial \theta_i}  -\frac{1}{2}\frac{\partial \log(2\pi)}{\partial \theta_i} \\
& = -\frac{1}{2} \tr\left(\mathcal{K}^{-1}\frac{\partial \mathcal{K}}{\partial \theta_i} \right) -\frac{1}{2} \left(\mathbf{y}^T\mathcal{K}^{-1}\frac{\partial \mathcal{K}}{\partial \theta_i}\mathcal{K}^{-1}\mathbf{y} \right)\\
& = -\frac{1}{2} \tr\left(\mathcal{K}^{-1}\frac{\partial \mathcal{K}}{\partial \theta_i} \right) -\frac{1}{2} \tr \left(\mathcal{K}^{-1}\mathbf{y}\mathbf{y}^T\mathcal{K}^{-1}\frac{\partial \mathcal{K}}{\partial \theta_i} \right) \\
& = \frac{1}{2} \tr\left( (\mathcal{K}^{-1}\mathbf{y}\mathbf{y}^T\mathcal{K}^{-1}-\mathcal{K}^{-1}) \frac{\partial \mathcal{K}}{\partial \theta_i}\right)
\end{split}
\label{eq:MLE_Appendix}
\end{equation} 
\end{lemma}

\begin{lemma}[Derivatives of noisy RBF Kernel w.r.t. $\theta \triangleq \protect{[\sigma_{sig},\sigma_n]} $]
\label{lemma:kernel_derivatives}
The derivatives of the RBF kernel with respect to its parameters are given by:

\begin{equation}
\frac{\partial \Sigma}{\partial \sigma^2_{sig}} = \exp \left( \frac{- ||\mathbf{x}-\mathbf{x}'||}{2}\right) = \mathcal{K}
\label{eq:dk_dsig}
\end{equation}


\begin{equation}
\frac{\partial \Sigma}{\partial \sigma^2_n} = I
\label{eq:dk_dnoise}
\end{equation}
\end{lemma}

\begin{theorem}[Maximum Likelihood Estimation]
\label{app:MLE}
From Eq.~\eqref{eq:logLikelihood}, we already know the expression for log-likelihood. We will just update that for the covariance kernel $\Sigma$ from Eq.~\eqref{eq:sq_exp_cov_hps}, such that:
\begin{equation}
\mathcal{LL} = \underbrace{-\dfrac{n}{2}\log(2\pi)-\dfrac{n}{2}\log(\textcolor{blue}{\sigma_{sig}}^2)}_\text{Const.}-\underbrace{\dfrac{1}{2}\log|\Sigma|}_\text{Complexity}-\underbrace{\dfrac{1}{2\textcolor{blue}{\sigma_{sig}}^2}Y^T\Sigma^{-1}Y}_\text{Data fit}
\label{eq:logLikelihood}
\end{equation}
Let $\hat{\tau}^2$ represent the scale of the data which is associated with the signal variance $\sigma_{sig}^2$ such that $\hat{\tau}^2 = 2 \sigma_{sig}$. Then, for the covariance kernel $\Sigma$ from Eq.~\eqref{eq:sq_exp_cov_hps}, we have the following relationships:

\begin{align}
\begin{split}
\hat{\sigma}_{sig} &= \dfrac{Y^T\Sigma^{-1}Y}{n}\,.\\   
\hat{\sigma}_{n} &= \dfrac{n}{2}\dfrac{Y^T(\Sigma^{-1})^2Y}{Y^T\Sigma^{-1}Y} - \dfrac{1}{2}\tr(\Sigma^{-1})
\end{split}    
\end{align}
\end{theorem}

\begin{proof}
First, we will find the derative of the log likelihood with respect to the signal standard deviation $\sigma_{sig}$ to get the optimal estimate $\hat{\sigma}_{sig}$. Thus,

$\dfrac{\partial \mathcal{LL}}{\partial \sigma_{sig}} = \dfrac{-n}{\sigma_{sig}} + \dfrac{Y^T\Sigma^{-1}Y}{\sigma_{sig}^3}$. 

Setting this gradient to zero to get the optimal estimate, we get:
\begin{equation}
\begin{split}
    \dfrac{-n}{\sigma_{sig}} + \dfrac{Y^T\Sigma^{-1}Y}{\sigma_{sig}^3} = 0 \\
    \Rightarrow \hat{\sigma}_{sig}^2 = \dfrac{Y^T\Sigma^{-1}Y}{n}\,.
\end{split}
\end{equation}

Now, we can plug back this optimal signal variance into the equation of log-likelihood to get
\begin{equation}
    \mathcal{LL}(\sigma_n) = \underbrace{-\dfrac{n}{2}\log(2\pi)-\dfrac{n}{2}\log(\dfrac{Y^T\Sigma^{-1}Y}{n})}_\text{Const.}-\underbrace{\dfrac{1}{2}\log|\Sigma|}_\text{Complexity}-\underbrace{\dfrac{1}{2\dfrac{Y^T\Sigma^{-1}Y}{n}}Y^T\Sigma^{-1}Y}_\text{Data fit}
\end{equation}

Now, in order to get the optimal noise variance, we simply need to set $\dfrac{\partial \mathcal{LL}(\sigma_n)}{\sigma_n} = 0$ which is purely a function of $\sigma_n$. Then, by utilizing the Lemma~\eqref{lemma:lml} and Lemma~\eqref{lemma:kernel_derivatives} we can obtain the optimal estimate of $\hat{\sigma}_n$.
\end{proof}

\subsection{Entropy of GP}
\label{app:entropy_GP}

\begin{lemma}[Symmetry of trace of product of $2$ matrices]
Suppose $P\in\mathcal{R}^{m\times n}$ and $Q\in\mathcal{R}^{n\times m}$, then $\tr(PQ) = \tr(QP)$.
\label{lemma:SymmTr}
\end{lemma}
\begin{proof}
By the definition of $\tr(\cdot)$, we know that \newline 
$tr(PQ) = \Sigma_{i=1}^m(PQ)_{ii} = \Sigma_{i=1}^m \Sigma_{j=1}^n P_{ij}Q_{ji}= \Sigma_{i=1}^n \Sigma_{j=1}^m Q_{ji}P_{ij}= \Sigma_{j=1}^m(QP)_{jj} = \tr(QP)$.
\end{proof}

\begin{corollary}[Symmetry of trace for $3$ matrices]
Suppose $P\in\mathcal{R}^{m\times n}$, $Q\in\mathcal{R}^{n\times o}$ and $R\in\mathcal{R}^{o\times m}$ then $\tr(PQR) = \tr(RQP)$.
\label{corr:SymmTr}
\end{corollary}

\begin{proof}
Let $S=PQ$ and $T=R$. Then from Lemma~\eqref{lemma:SymmTr}, we already know that:
\begin{align}
\begin{split}
\tr(ST) = \tr(TS) \\
\implies \tr(PQR) = \tr(RQP)
\end{split}
\label{eq:SymmTr}
\end{align}

\end{proof}

\begin{theorem}[Entropy of GP]
Let $\mathcal{K}_{f|D}$ represent the posterior covariance of a GP for set $X$ representing the observed inputs and set $X^*$ standing for unobserved inputs. Let $Y^{*}$ represent the measurements for $X^*$ and the training data $D = [X,Y]$. Also, say $\#(X^*)$ defines the cardinality of the $X^*$. Then, the conditional entropy $\mathcal{H}[Y^*|D]$ is denoted by:
\begin{equation}
\mathcal{H}[Y^*|D] = \frac{1}{2}\log[(2\pi e)^{\#(X^*)}|\mathcal{K}_{f|D}|]
\end{equation}
\end{theorem}

\begin{proof}
Consider a column vector of random measurements $Y^*$ for inputs belonging to set $X^*$. We know that $Y^* \sim \mathcal{N}(\bm{\mu}_{f|D},\mathcal{K}_{f|D})$ with a pdf given by:
\begin{equation}
\psi(Y^*)= \frac{1}{\sqrt{(2\pi)^{\#(X^*)}|\mathcal{K}_{f|D}|}}
\exp\left(-\frac{1}{2}(Y^*-{\bm{\mu}_{f|D}})^T{\mathcal{K}_{f|D}}^{-1}(Y^*-{\bm{\mu}_{f|D}})
\right)
\end{equation}
Then, by the definition of Shannon entropy over the continuous domain, we have
\begin{align}
\begin{split}
&\mathcal{H}[Y^*|D] \\
&= -\int\left\lbrace \psi(Y^*) \log(\psi(Y^*)) \right\rbrace dY^* \\
&= -\int\left\lbrace \psi(Y^*) \left[-\frac{1}{2}(Y^*-\bm{\mu}_{f|D})^T \mathcal{K}_{f|D}^{-1}(Y^*-\bm{\mu}_{f|D}) - \log(\sqrt{(2\pi)^{\#(U)}|\mathcal{K}_{f|D}|})\right] \right\rbrace dY^* \\
&= \frac{1}{2}\E_\psi[\tr \left\lbrace(Y^*-\bm{\mu}_{f|D})^T \mathcal{K}_{f|D}^{-1}(Y^*-\bm{\mu}_{f|D})\right\rbrace] + \frac{1}{2} \log\left[(2\pi)^{\#(X^*)}|\mathcal{K}_{f|D}|\right]\\
&\text{using Corollary~\eqref{corr:SymmTr}, we get:}\\
&=\frac{1}{2}\E_\psi[\tr \left\lbrace(Y^*-\bm{\mu}_{f|D})(Y^*-\bm{\mu}_{f|D})^T \mathcal{K}_{f|D}^{-1}\right\rbrace] + \frac{1}{2} \log\left[(2\pi)^{\#(X^*)}|\mathcal{K}_{f|D}|\right]\\
&=\frac{1}{2}\E_\psi[\tr \left\lbrace(Y^*-\bm{\mu}_{f|D})(Y^*-\bm{\mu}_{f|D})^T\right\rbrace]\mathcal{K}_{f|D}^{-1} + \frac{1}{2} \log\left[(2\pi)^{\#(X^*)}|\mathcal{K}_{f|D}|\right]\\
&=\frac{1}{2}\tr \left\lbrace \mathcal{K}_{f|D}\mathcal{K}_{f|D}^{-1}\right\rbrace + \frac{1}{2} \log\left[(2\pi)^{\#(X^*)}|\mathcal{K}_{f|D}|\right]\\
&=\frac{1}{2}\tr \left\lbrace \mathcal{I} \right\rbrace + \frac{1}{2} \log\left[(2\pi)^{\#(X^*)}|\mathcal{K}_{f|D}|\right]\\
&=\frac{1}{2}\#(X^*) + \frac{1}{2} \log\left[(2\pi)^{\#(U)}|\mathcal{K}_{f|D}|\right]\\
&= \frac{1}{2} \log\left[(2\pi e)^{\#(U)}|\mathcal{K}_{f|D}|\right]
\end{split}
\end{align}

\end{proof}
\cleardoublepage
\section*{Bibliography}
\bibliographystyle{model1-num-names}
\bibliography{ref.bib}

\end{document}